\documentclass[3p,times]{elsarticle}
\usepackage{amssymb}
\usepackage{colortbl}
\usepackage[colorlinks=true, urlcolor=blue]{hyperref}
\usepackage{lineno}
\usepackage{amsmath}
\usepackage{graphicx}
\usepackage{todonotes}
\usepackage{multirow}
\usepackage{booktabs}
\usepackage{tikz}
\usetikzlibrary{positioning,arrows.meta,calc,matrix}

\usepackage{xcolor}
\definecolor{stepone}{RGB}{0,90,160}
\definecolor{steptwo}{RGB}{0,128,128}
\definecolor{stepthree}{RGB}{120,70,170}
\definecolor{stepfour}{RGB}{220,120,20}

\newcommand{\StepOne}[1]{\textcolor{stepone}{#1}}
\newcommand{\StepTwo}[1]{\textcolor{steptwo}{#1}}
\newcommand{\StepThree}[1]{\textcolor{stepthree}{#1}}
\newcommand{\StepFour}[1]{\textcolor{stepfour}{#1}}
\usepackage{soulutf8}

\definecolor{myyellowtext}{RGB}{255,215,0}
\renewcommand{\hl}[1]{\textcolor{black}{#1}}
\begin{document}
\begin{frontmatter}

\title{An Uncertainty Estimation Framework for Dose Accumulation in Adaptive Radiotherapy: Application to CBCT-Guided Radiotherapy for Cervical Cancer}

\author{Cédric Hémon\corref{cor1}$^{1\dagger}$, Delphine Lebret$^{1\dagger}$, Jean-Claude Nunes$^1$, Valentin Boussot$^1$, Karine Peignaux$^2$, Nathalie Mesgouez-Nebout$^3$, Chantal Hanzen$^4$, Antoine Simon$^1$, Anaïs Barateau$^1$, Renaud de Crevoisier$^1$, Caroline Lafond$^1$}

\cortext[cor1]{Corresponding author}
\ead{cedric.hemon@univ-rennes.fr}
\address{$^1$ Univ. Rennes, CLCC Eug\`ene Marquis, INSERM, LTSI - UMR 1099, F-35000 Rennes, France \\
$^2$ Department of Radiation Oncology, Centre Georges-Francois Leclerc, F-21000 Dijon, France\\
$^3$ Institut de Cancérologie de l'Ouest–Site Paul Papin, F-49000 Angers, France\\
$^4$ CLCC Henri Becquerel, Rouen, France, F-76000 Rouen, France\\
$^\dagger$ These authors contributed equally to this work.
}
\begin{abstract}
\textbf {Background and purpose:} Online adaptive radiotherapy (oART) enables daily treatment plan adaptation to account for interfraction anatomical variations. However, accurate cumulative dose estimation remains limited by uncertainties in deformable image registration (DIR), segmentation, and anatomical variability. We introduce IMPACT-DoseAcc, an uncertainty-aware dose accumulation framework integrated within the broader IMPACT ecosystem for semantic feature–driven multimodal medical image analysis in radiotherapy. The framework is designed to be modality- and disease-agnostic, and is here applied to cone-beam CT (CBCT)–guided oART for locally advanced cervical cancer (LACC).\\
\textbf {Material and Methods:} Nine patients with LACC were retrospectively analyzed using daily CBCT-derived virtual CTs (vCTs) for dose recalculation. In this work, IMPACT-DoseAcc focuses exclusively on predictive uncertainty arising from DIR, without modeling uncertainties related to vCT generation. Two complementary DIR uncertainty estimation strategies were investigated, both leveraging deep segmentation feature representations within the IMPACT-Reg framework, but differing in the choice of the feature extraction model: (1) a Bayesian segmentation-guided approach based on a single probabilistic model to quantify anatomical uncertainty, and (2) an ensemble of independently trained segmentation models targeting distinct anatomical structures to capture epistemic variability across feature representations. Voxel-wise uncertainty maps were propagated through dose warping and accumulation to generate probabilistic dose–volume histograms (pDVHs). \hl{Ensemble-based DIR uncertainty was quantified from the voxel-wise standard deviation across deformation fields, and geometric error was evaluated using the average surface distance (ASD) between warped and clinician-validated contours.} An anatomical-variability–based weighting scheme was introduced to refine cross-fraction uncertainty aggregation. All components were implemented as modular plugins in 3DSlicer to ensure reproducibility and clinical usability.\\
\textbf{Results: } Ensemble-based DIR uncertainty showed strong correlation with geometric error (mean Pearson correlation coefficient = 0.63 for CTVt and 0.66 for the bladder). For the CTVt, pDVHs achieved 96.3 ± 3.9\% coverage, demonstrating appropriate calibration of propagated uncertainties. Anatomical-variability weighting further stabilized uncertainty estimates across fractions and organs.\\
\textbf {Conclusions:} IMPACT-DoseAcc enables voxel-wise propagation of registration-driven uncertainty to cumulative dose metrics, improving interpretation of accumulated dose under realistic anatomical variations. Its integration within 3DSlicer supports reproducible and uncertainty-informed ART workflows.
\end{abstract}

\begin{keyword}
Dose accumulation \sep Online adaptive radiotherapy \sep Cervical cancer \sep Deformable image registration \sep Uncertainty quantification \sep 3DSlicer
\end{keyword}

\end{frontmatter}

\section{Introduction}

Locally advanced cervical cancer (LACC) radiotherapy presents significant clinical challenges due to substantial inter- and intra-fractional anatomical variations. \hl{Early studies have shown that anatomical changes during cervical radiotherapy may lead to discrepancies between planned and delivered pelvic dose, as well as significant interfraction dose variations during intensity-modulated radiotherapy based on weekly CT evaluation~\cite{lim_pelvic_2009,han_interfractional_2006}. More recent work further investigated the dosimetric impact of pelvic organ motion on target volumes and evaluated the effect of bladder filling on daily dose distributions~\cite{eminowicz_understanding_2017,zhang_evaluation_2023}.} These variations increase the risk of tumor underdosage or organ-at-risk (OAR) overexposure, potentially compromising both treatment efficacy and patient safety. Adaptive radiotherapy (ART) has therefore emerged as a strategy to dynamically adjust treatment according to the patient’s anatomy at each session.

ART strategies, whether based on predefined plan libraries or daily online re-optimization, must account for pronounced pelvic organ motion and deformation throughout treatment. \hl{Adaptive strategies based on pretreatment anatomical scenarios or daily plan selection have already been investigated \cite{Bondar2012-PoD,Heijkoop2014-PoD,Van_de_Schoot2017-PoD}, while Rigaud et al.}~\cite{Rigaud2018-PoD}\hl{~and Lebret et al.}~\cite{Lebret2025ARCOL}\hl{~reported cervical cancer implementations of library-based or plan-of-the-day workflows. More recently, Ding et al.}~\cite{Ding2025online}\hl{~described an MRI-guided daily online adaptive strategy.} The uterus, cervix, bladder, rectum, and bowel loops may shift or change shape between and within fractions, leading to discrepancies between planned and delivered dose distributions. Accurate estimation of the cumulative delivered dose is therefore essential to assess the true dosimetric impact of anatomical variations and to evaluate the clinical benefit of adaptive strategies. In current CBCT-guided workflows, cumulative dose estimation typically relies on synthetic or virtual CTs (sCTs/vCTs) generated from cone-beam CT (CBCT) for daily dose recalculation. \hl{In this manuscript, sCT is used as a generic term for CT-like images synthesized from another modality, whereas vCT specifically refers to the RayStation sCT generation method \cite{weistrand2015anaconda} used in our workflow.}

\hl{Dose accumulation combines fraction-wise dose distributions into a common reference frame, usually the planning CT (pCT), through deformable image registration (DIR) and associated deformation vector fields (DVFs), as described by Chetty and Rosu-Bubulac}~\cite{chetty2019deformable} \hl{and by McDonald et al.}~\cite{mcdonald2023dose}\hl{. However, uncertainties inherent to DIR and related processes remain a major limitation for reliable cumulative dose assessment. The AAPM TG-132 report by Brock et al.}~\cite{brock_tg132_2017} \hl{emphasized that registration accuracy must be assessed in relation to the intended clinical application, and Paganelli et al.}~\cite{paganelli_patient-specific_2018} \hl{reviewed patient-specific validation strategies and their limitations in radiotherapy. More recently, Nenoff et al.}~\cite{nenoff2023review} \hl{provided radiotherapy-specific recommendations for DIR uncertainty evaluation, while Murr et al.}~\cite{murr2023applicability} \hl{discussed the particular implications of uncertainty for dose mapping and accumulation. In this context, no single uncertainty metric is considered sufficient across all use cases, especially when geometric and dosimetric criteria need to be interpreted jointly, as also discussed by H\'emon et al.}~\cite{hemon2026towards}\hl{.} \hl{CBCT-based workflows are particularly prone to uncertainty due to limited soft-tissue contrast, imaging artifacts, and inter-modality inconsistencies between CBCT and CT, which may affect registration robustness, and intensity variability across multislice CT and CBCT has been documented by Nackaerts et al.}~\cite{nackaerts2011analysis}\hl{. Bayesian and probabilistic uncertainty quantification (UQ) approaches have been explored in monomodal settings by Cui et al.}~\cite{cui_bayesian_2021}\hl{, Khawaled et al.}~\cite{khawaled_npbdreg_2022}\hl{, Grzech et al.}~\cite{grzech_variational_2022}\hl{, Hua et al.}~\cite{hua_variational_2024}\hl{, and Chen et al.}~\cite{chen_registration_2024}\hl{, whereas Nenoff et al.}~\cite{nenoff_deformable_2020}\hl{~and Smolders et al.}~\cite{smolders2022deformable}\hl{~highlighted the dosimetric relevance of DIR uncertainty for dose accumulation. Its application to multimodal registration and downstream dose accumulation nevertheless remains limited, with radiotherapy-focused examples discussed by Meyer et al.}~\cite{MEYER2025}\hl{. In high dose-gradient regions near sensitive organs, even small registration errors may translate into clinically relevant dosimetric deviations, as illustrated by Nenoff et al.}~\cite{nenoff_deformable_2020}\hl{~and Veiga et al.}~\cite{veiga2015toward}\hl{.}

\hl{Beyond image registration, uncertainty arises at multiple stages of the adaptive workflow. In particular, sCT generation from CBCT has been shown to introduce uncertainty in voxel-wise dose computation due to image synthesis variability and inaccuracies in Hounsfield unit estimation~\cite{quintero2025evaluation,hemon2025modeling}. Similarly, segmentation variability, arising either from inter-operator differences in manual delineation or from uncertainty in automatic segmentation algorithms, may propagate through registration and dose warping, influencing both geometric alignment and DVH metrics. Importantly, such variability can also impact the dose planning stage itself, as contour definition directly affects target volumes and organs at risk delineation. Although recent studies have begun to explore uncertainty propagation across cascaded medical imaging tasks, notably by Mehta et al.}~\cite{mehta2021propagating}\hl{~and Feiner et al.}~\cite{feiner2023propagation}\hl{, a unified framework capable of consistently aggregating uncertainties from image synthesis, segmentation, registration, and dose accumulation is still lacking.}

To address these challenges, we introduce IMPACT-DoseAcc, an uncertainty-aware dose accumulation framework for online ART (oART), applied here to CBCT-guided ART for LACC. Developed within the broader IMPACT ecosystem for semantic feature-driven multimodal image analysis, IMPACT-DoseAcc builds upon the IMPACT-Reg framework proposed by Boussot et al.~\cite{boussot_impact_2025}. \hl{It enables voxel-wise propagation of registration- and anatomy-related uncertainties through dose warping and accumulation. This results in probabilistic dose-volume histograms (pDVHs) and spatial uncertainty maps.} The ecosystem (IMPACT-Reg, IMPACT-Synth, and IMPACT-DoseAcc) is fully integrated into \href{https://github.com/vboussot/SlicerImpactDoseAcc}{3DSlicer}, providing a reproducible and clinically oriented environment for ART support.

\hl{Although IMPACT-DoseAcc is designed as a general framework for uncertainty-aware dose accumulation, the present study focuses specifically on uncertainty arising from DIR, because it remains a dominant contributor to variability in cumulative dose estimation~\cite{chetty2019deformable,hemon2026towards}. The integration of synthesis- and segmentation-related uncertainties into a comprehensive probabilistic accumulation framework constitutes an important direction for future work. Accordingly, the present study should be regarded as a proof-of-concept evaluation of registration-driven uncertainty propagation in a limited multi-institutional longitudinal cohort of patients with LACC.}

We hypothesize that inaccuracies in anatomical correspondence due to registration errors manifest as increased variability among deformation fields generated by different models. This inter-model variability can be quantified using an ensemble-based strategy to derive voxel-wise uncertainty maps reflecting spatial confidence in the estimated transformations. Such information is critical for identifying regions where deformation uncertainty may compromise dosimetric accuracy and clinical decision-making.

The main contributions of this study are:
\begin{itemize}
    \item \textbf{An end-to-end CBCT-based oART framework} incorporating vCT generation from daily CBCTs using the \hl{ANACONDA} deformable registration algorithm implemented in RayStation (RaySearch Laboratories, Sweden) \cite{weistrand2015anaconda}, followed by daily online plan re-optimization on the anatomy of the day.
    \item \textbf{A novel uncertainty-aware DIR approach} based on the IMPACT metric within IMPACT-Reg \cite{boussot_impact_2025}, leveraging semantic features from multiple pretrained models to estimate registration uncertainty in multimodal settings.
    \item \textbf{A Bayesian adaptation of the TotalSegmentator model} \cite{wasserthal2023totalsegmentator} enabling estimation of anatomical uncertainty within a probabilistic segmentation framework.
    \item \textbf{Voxel-wise uncertainty propagation} throughout the dose accumulation process, allowing confidence intervals to be derived for DVHs and cumulative dose maps.
\end{itemize}

This work aims to enhance clinical decision support, treatment adaptation, and quality assurance in uncertainty-aware ART.

\section{Material and Methods}

\subsection{Dataset}

This retrospective study included nine patients treated with external beam radiotherapy for LACC between 2017 and 2022 across four French care centers. The prescribed dose was 45~Gy delivered in 25 fractions of 1.8~Gy each~\cite{Lebret2025ARCOL}. 

Planning CTs, whose characteristics are summarized in Table~\ref{tab:infosCTCBCT}, were manually delineated by experienced radiation oncologists following the EMBRACE~II clinical protocol~\cite{potter2018embrace}. These contours served as reference structures for subsequent registration and dose accumulation analyses.

For each treatment fraction, daily CBCT scans, as detailed in Table~\ref{tab:infosCTCBCT}, were acquired throughout treatment to capture inter-fraction anatomical variations. CBCTs were delineated using a deep learning–based segmentation algorithm specifically trained on pelvic anatomy \cite{isensee2020nnu}. \hl{The resulting contours were visually inspected and validated by an expert radiation oncologist to ensure anatomical and clinical consistency as described in \cite{Lebret2025ARCOL}. They were then directly transferred onto the corresponding vCT images without any deformation.}

\begin{table}[h!]
\centering
\begin{tabular}{lllllllll}
\toprule
\textbf{Patient} & \textbf{Center} & \textbf{CT system} & \textbf{CT size (mm)} & \textbf{CBCT system} & \textbf{kV} & \textbf{mAs} & \textbf{CBCT size (mm)} \\
\midrule
1 & 1 & Big Bore, Philips   & 600,600,462   & XVI, Elekta  &120 &1000 & 410,410,264 \\
2 & 1 & Somatom, Siemens  & 650,650,444   & XVI, Elekta &120 &1000  & 410,410,264  \\
3 & 2 & Lightspeed, GE Healthcare  & 650,650,395   & OBI, Varian &125 &1082   & 465,465,175   \\
4 & 2 & Lightspeed, GE Healthcare  & 650,650,440   & OBI, Varian &125 &1082   & 465,465,175 \\
5 & 3 & PatXferRT, Brainlab & 650,650,365   & OBI, Varian &125 &1074    & 465,465,175   \\
6 & 3 & PatXferRT, Brainlab & 650,650,528   & OBI, Varian &125 &1074   & 465,465,175   \\
7 & 4 & Aria, Varian & 537,537,443   & OBI, Varian &125 &1082   & 465,465,175   \\
8 & 4 & Optima, GE Healthcare & 511,511,343  & OBI, Varian &125 &700   & 450,450,160 \\
9 & 4 & Optima, GE Healthcare  & 500,500,515    & OBI, Varian &125 &709   & 450,450,160   \\
\bottomrule
\end{tabular}
\caption{CT and CBCT imaging characteristics.} 
\label{tab:infosCTCBCT}
\end{table}

\subsection{Online Adaptive Radiotherapy}

An oART strategy was implemented and optimized on the anatomy of the day using vCTs. An isotropic 5 mm CTV-to-PTV margin was applied to account for intra-fraction anatomical changes, consistent with previously reported values for both CTVt and CTVn across patients and treatment durations \cite{yen2024improved,wang2024assessing}.

Daily adaptive plans were simulated using VMAT on a VersaHD (Elekta) linear accelerator. The isocenter was positioned at the center of the PTV. Treatments were delivered with 6 MV flattening-filter-free beams using two coplanar arcs.

Daily plan optimization was performed directly on the vCT using the RayStation (RaySearch Laboratories, Sweden) treatment planning system, maintaining identical dosimetric objectives and constraints as in the initial plan (from the pCT). 
 

\subsection{Dose Accumulation}
\begin{figure}[h!]
    \centering
    \includegraphics[width=\textwidth]{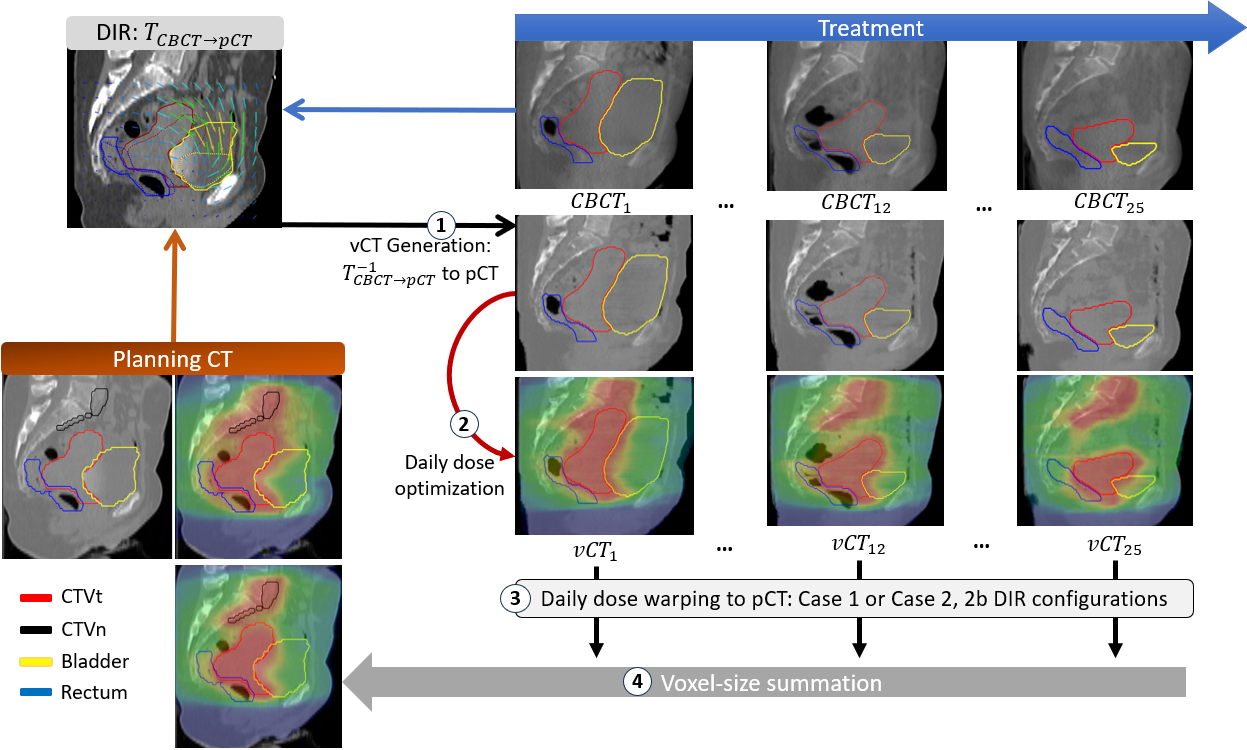}
    \caption{Study workflow: (1) Daily CBCT images were converted into vCTs using ANACONDA DIR with the pCT, (2) the daily dose was then re-optimized on each vCT, (3) DVFs between vCT and pCT were computed to warp the daily doses to the pCT space, and (4) voxel-wise dose accumulation was performed on the pCT. 
    \\
    Abbreviations: CBCT = cone-beam computed tomography; vCT = virtual computed tomography; pCT = planning computed tomography; DIR = deformable image registration.}
    \label{fig:workflow}
\end{figure}

In this study, dose accumulation was carried out using a multi-step pipeline (Fig.\ref{fig:workflow}) integrating onboard imaging, vCT generation, DIR, and voxel-wise dose warping. This process was designed to capture the cumulative dose delivered throughout treatment by accounting for anatomical variations observed across fractions.

The dose accumulation procedure was as follows (Figure~\ref{fig:workflow}):
\begin{itemize} 
\item \textbf{vCT Generation:} Each CBCT scan was converted into a corresponding vCT using the \hl{ANACONDA} DIR algorithm \cite{weistrand2015anaconda} implemented in RayStation. \hl{This algorithm performs deformable registration with the daily CBCT as the target image and the pCT as the reference image, initialized by a rigid alignment between the pCT and CBCT. The pCT is then deformed to match the patient’s anatomy of the day as depicted on the CBCT, while preserving its original Hounsfield unit values for accurate dose calculation. Outside the CBCT field of view (FOV), anatomical information from the pCT is propagated to complete the vCT. A contour-guided registration mode was employed to regularize the deformable step and limit boundary artifacts in clinically relevant pelvic regions, using the CTVt, bladder, and rectum as guiding structures.}
\item \textbf{Daily Dose Optimization:} Using the vCTs, the dose for each fraction was reoptimized based on the actual daily anatomy.
\item \textbf{DIR:} Our uncertainty-aware DIR approaches were employed to generate DVFs between the vCT of each fraction and the pCT.
\item \textbf{Dose Warping and Accumulation:} The dose distributions of the vCTs were warped to the pCT coordinate space using the corresponding DVFs. A voxel-wise summation of all warped doses was then performed.
\end{itemize}

The resulting cumulative dose map in the reference space of the pCT enabled direct voxel-wise comparison with the initially prescribed dose distribution. \hl{This workflow provided a pragmatic estimate of the delivered dose because dose recalculation was performed on anatomy-of-the-day vCTs and each daily dose was subsequently mapped back to a common pCT reference frame before accumulation. However, the resulting accumulated dose remains conditioned on the geometric fidelity of the underlying vCTs and should therefore be interpreted as a surrogate of delivered dose rather than as an absolute ground truth.} This allowed local assessment of potential under- or overexposure in tumor volumes and OARs.

\subsection{Uncertainty Quantification of Registration for Dose Warping}

We propose two complementary approaches to uncertainty estimation in DIR. The first involves structure-guided uncertainty by integrating probabilistic segmentation models into the registration process. The second employs ensembles of feature extraction networks within the similarity metric, thus capturing model variability and ambiguity in learned representations. Both strategies aim to produce spatial uncertainty maps that highlight regions of low registration confidence.

These approaches build upon our previously proposed IMPACT similarity measure of IMPACT-Reg \cite{boussot_impact_2025}, which leverages semantic features from pre-trained and foundation models to guide the registration process. The IMPACT-based method has been integrated into the conventional \href{https://github.com/vboussot/ImpactLoss}{Elastix framework} \cite{klein2009elastix}, enabling robust and uncertainty-aware DIR. The effectiveness of the multimodal registration approach has been systematically assessed in multiple international \href{https://learn2reg.grand-challenge.org/}{Learn2Reg challenges}, including ThoraxCBCT, AbdomenMRCT, and LungCT, where it achieved top-ranking performance \cite{boussot_impact_2025} and demonstrated strong generalization across anatomically and modality-diverse tasks.

\hl{In the present work, this validated registration backbone is used not only to generate DVFs, but also to investigate how DIR uncertainty propagates through the complete dose-accumulation workflow. The analysis was organized as a four-step evaluation pipeline: all candidate DIR configurations are first compared geometrically, the retained ensemble-based uncertainty model is then validated against geometric error, the main uncertainty models are next compared in the dose domain, and the selected aggregation strategy is finally interpreted through clinical and temporal/spatial analyses.}

For all subsequent geometric and dosimetric evaluations, three DIR configurations  (Figure~\ref{fig:cases_aggregation_overview}) were derived from these strategies:
\begin{itemize}
    \item \textbf{Case 1}: an ensemble-based approach combining the DVFs generated from eight heterogeneous deep models.
    \item \textbf{Case 2a}: a Bayesian TotalSegmentator model \cite{wasserthal2023totalsegmentator} fine-tuned with additional labels for the peritoneal cavity and cervix, enabling the propagation of segmentation-driven anatomical uncertainty.
    \item \textbf{Case 2b}: a corresponding Bayesian model fine-tuned without the peritoneal cavity label.
\end{itemize}

\hl{Accordingly, the study was structured into four investigation steps (Fig.~\ref{fig:cases_aggregation_overview}):}
\begin{itemize}
    \item \hl{Step 1 - Configuration comparison and geometric validation: evaluation of the individual feature sources together with the three DIR configurations (Case 1, Case 2a, and Case 2b).}
    \item \hl{Step 2 - Uncertainty validation: assessment of whether the ensemble-based uncertainty estimate of Case~1 is spatially consistent with contour-propagation error measured by ASD.}
    \item \hl{Step 3 - Dose-domain comparison: comparison of accumulated dose and pDVHs for Case~1 and Case~2 against reference DVHs.}
    \item \hl{Step 4 - Aggregation and interpretation: comparison of quadratic and bladder-similarity weighted aggregation, followed by clinical, temporal/spatial, and visualization analyses of the selected strategy.}
\end{itemize}

These three DIR configurations (Case 1, Case 2a, and Case 2b) were used consistently throughout the study.

\subsubsection{Bayesian segmentation model for DIR uncertainty estimation}

This Bayesian segmentation-guided approach leverages the nature of deep segmentation models \cite{riera2025calibration}. Probabilistic segmentation networks output multiple plausible segmentations instead of a single deterministic output, providing insight into anatomical-structure variability. By incorporating this information into the segmentation-guided registration process, we can estimate the uncertainty of the resulting spatial transformation.

This method for estimating Bayesian uncertainty \cite{riera2025calibration} was evaluated as part of the \href{https://curvas.grand-challenge.org/}{CURVAS MICCAI challenge} , where it achieved third place overall, and notably ranked first in terms of calibration of uncertainty, demonstrating its ability to provide both accurate and well-calibrated probabilistic segmentations. Further implementation details and the strategy used to enable UQ via minimal fine-tuning are described below.

Let $G_{prob}$ be a probabilistic segmentation network applied to two rigid registered images, $\mathbf{I}_A$ and $\mathbf{I}_B$. Instead of providing a single segmentation output, $G_{prob}$ generates multiple segmentation samples $\mathbf{S}_{A}^{(i)}$ and $\mathbf{S}_{B}^{(i)}$ using Monte Carlo (MC) sampling:

\begin{equation}
\mathbf{S}_{A}^{(i)} = G_{prob}(\mathbf{I}_A, \theta_i), \quad \mathbf{S}_{B}^{(i)} = G_{prob}(\mathbf{I}_B, \theta_i), \quad i = 1, \dots, N
\end{equation}

where $\theta_i$ represents different realizations of the network's weights obtained through variational inference.

The  $\mathbf{I}_A/\mathbf{I}_B$ deformable registration process is then performed on each sampled segmentation map to compute a set of  DVFs $\Phi^{(i)}$:

\begin{equation}
\Phi^{(i)} = \arg\min_{\Phi} \mathcal{S}(\mathbf{S}_{A}^{(i)}, \mathbf{S}_{B}^{(i)}),
\end{equation}

where $\mathcal{S}$ represents the similarity metric used for $\mathbf{I}_A/\mathbf{I}_B$ registration. In this implementation, the feature maps from the last decoder layer are compared using the L1 norm.

This method provides a data-driven approach to quantifying uncertainty in segmentation-guided registration, as it accounts for anatomical variability inherent in medical images.

\paragraph{\textbf{Fine-tuning network}}
\label{fine tuning}

Bayesian models \cite{hemon2026towards} are known for their ability to provide both accurate predictions and reliable uncertainty estimates. However, training such models from scratch can be unstable, especially with limited annotated 3D datasets, which may lead to poor generalization. To address this, we adopt a hybrid strategy: leveraging a pre-trained segmentation model while enabling uncertainty estimation through minimal fine-tuning.

We selected TotalSegmentator \cite{wasserthal2023totalsegmentator}, a state-of-the-art segmentation model based on five nnU-Net architectures trained on a large CT dataset. To transform this deterministic model into a Bayesian one, we used the Adaptable Bayesian Neural Network (ABNN) strategy \cite{franchi2024make}. Specifically, we replaced the original layer normalization layers with Bayesian Normalization Layers (BNL), introducing Gaussian perturbations to the normalization parameters. This allowed us to approximate a posterior distribution over the model's predictions with minimal architectural changes.
The BNL replacing standard layer normalization in the UNet architecture is defined as follows:
\begin{equation} 
\text{BNL}(h_j) = \frac{h_j - \hat{\mu}_{j}}{\hat{\sigma}_{j}} \times \gamma_j (1 + \epsilon_j) + \beta_j. \label{eq} 
\end{equation}

BNL$(h_j)$ represents an affine function with respect to the normalization layer (NL) of the layer $j$. Our normalization layers consist of trainable parameters $\gamma_j$ and $\beta_j$ (respectively scale and shift) of the layer $j$. The parameters $\hat{\mu}_j$ and $\hat{\sigma}_j$ denote the empirical mean and variance, respectively, of the layer $j$. The term $\epsilon \sim\mathbb{N}(0,1)$ refers to a reduced centered Gaussian perturbation applied to these parameters. These Gaussian perturbations are applied to the normalization layer weights before training begins, effectively introducing randomness that allows the model to approximate the posterior distribution of the weights.\\ 

During fine-tuning, only the scale and shift parameters of the BNL layers ($\gamma$ and $\beta$) were updated, while all other weights remained frozen. The model was retrained eight times for ten epochs each using an external dataset of 20 patients with LACC. Training followed the inference constraints and optimization scheme of the original TotalSegmentator model \cite{wasserthal2023totalsegmentator}, including its loss function, optimizer, and preprocessing pipeline. This empirical approach enabled uncertainty estimation across segmentation outputs without requiring full Bayesian inference over all network weights.

To further improve anatomical relevance, we fine-tuned the M291 TotalSegmentator model to incorporate an additional cervix label (Case 2b). An additional variant of the model was also trained that includes both the cervix and the peritoneal cavity as new labels (Case 2a). This model, originally trained to segment key pelvic structures such as the bladder, colon, and small bowel, was extended to enhance delineation specificity in critical regions for cervical cancer radiotherapy. The resulting probabilistic segmentation networks were integrated into the IMPACT metric within the IMPACT-Reg framework \cite{boussot_impact_2025}, enabling the propagation of anatomical uncertainty into the DIR process.

\subsubsection{Model ensemble for DIR uncertainty estimation}

An alternative strategy for estimating DIR uncertainty involves leveraging an ensemble of independently feature extraction models to extract the features that guide the registration. Rather than relying on a single model, this approach aggregates the outputs of multiple models, each potentially incorporating distinct prior knowledge or feature extraction capabilities derived from different pretraining objectives. Registration is then performed by optimizing a common similarity criterion computed in the corresponding feature spaces.

Each segmentation model in the ensemble contributes a unique "view" of the anatomical structures, informed by the specific task it was originally trained for. For example, we can consider that:
\begin{itemize} 
    \item models trained for \textit{organ segmentation} tend to emphasize macroscopic anatomical boundaries, 
    \item models specialized in \textit{vertebrae segmentation} capture rigid structures and fine spatial gradients, 
    \item models optimized for \textit{muscle segmentation} may be more sensitive to soft tissue contrast and variability. 
\end{itemize}

These task-specific perspectives add complementary information into the deformation estimation process. By performing registration using each of these DL-based segmentation models and analyzing the set of resulting deformation fields ${\Phi^{(1)}, \Phi^{(2)}, \dots, \Phi^{(M)}}$, one can assess the level of agreement across models. Areas where the deformations are consistent indicate high confidence in the registration output, whereas regions with substantial variability among the models suggest uncertainty due to ambiguity or lack of anatomical correspondence.

The registration uncertainty can be quantified by computing the voxel-wise variance across the ensemble of deformation fields:

\begin{equation} 
    U_{\text{ensemble}} = \frac{1}{M} \sum_{j=1}^{M} |\Phi^{(j)} - \bar{\Phi} |^2, \quad \text{with } \bar{\Phi} = \frac{1}{M} \sum_{j=1}^{M} \Phi^{(j)}, 
\end{equation}

where $M$ is the number of models in the ensemble and $\bar{\Phi}$ is the mean deformation field.\\
In this study, the ensemble included seven diverse models: TotalSegmentator \cite{wasserthal2023totalsegmentator} variants M291, M292, M294, M730, and M731; the MIND loss \cite{heinrich2012mind} (using a patch radius of 1 and dilation of 2); and the SAM encoder \cite{kirillov2023segment}. These models were selected to capture a broad spectrum of anatomical and structural features. The TotalSegmentator variants provide pretrained representations focused on different anatomical targets (e.g., organs, bones, soft tissues), while the MIND-based model emphasizes local structural similarity using modality-independent features. The SAM encoder \cite{kirillov2023segment} contributes high-level contextual information through a transformer-based architecture. Together, this heterogeneous ensemble enhances the robustness of feature extraction for registration and supports a richer estimation of epistemic uncertainty.

\subsection{Dose Accumulation Uncertainty}
In this study, we quantify dosimetric uncertainty during dose accumulation (Step 4) at the fraction level using two complementary approaches. 
First, uncertainty arising from the registration process is estimated by combining the dose variations obtained from multiple deformation models using a quadratic summation. 
Second, inter-fraction anatomical variability is explicitly accounted for by weighting the contribution of each treatment fraction based on bladder volume changes observed in daily imaging relative to the pCT.

\subsection*{Weighting Based on Inter-fraction Anatomical Variability}

To reflect inter-fraction anatomical variation, each fraction \( f \) is assigned a weighting factor \( w_f \) based on the Dice similarity coefficient ($Dice_f$) between the bladder segmentation on the pCT and the \hl{corresponding daily vCT}:
\begin{equation}
    w_f = 2\times(1-\text{Dice}_f),
\end{equation}
Fractions with high anatomical similarity (i.e., Dice close to 1) are assigned lower weights, while those with lower similarity (indicating larger anatomical deviations) contribute more significantly to the cumulative uncertainty. \hl{The bladder was selected because it undergoes large day-to-day volume variations in LACC radiotherapy \cite{TAYLOR2008250} and remains visible and quantifiable on CBCT images, making it a practical surrogate of global pelvic anatomical change.} 

\subsection*{Voxel-wise Weighted Uncertainty per Fraction}

For each treatment fraction $f$, the dose distribution \( D_f[\mathrm{Gy}] \) is first computed on the corresponding daily CT (or vCT), and then mapped to the pCT coordinate system using the DVF \( \mathbf{T}_f^{(k)} \) associated with registration model \( k \):
\begin{equation}
D_f^{(k) \rightarrow \mathrm{pCT}}(\mathbf{x}) = \mathbf{T}_f^{(k)}(D_f(\mathbf{x}))
\end{equation}

To capture model-related uncertainty at the fraction level, voxel-wise statistics are computed across the ensemble of \( K \) registration models for each fraction \( f \). Specifically, for a given voxel \( \mathbf{x} \), we compute the mean $\bar{D}_f(\mathbf{x})$ and standard deviation $\sigma_f(\mathbf{x})$ of the warped dose maps across models:
\begin{equation}
\sigma_f(\mathbf{x}) = \sqrt{ \frac{1}{K} \sum_{k=1}^{K} \left( D_f^{(k) \rightarrow \mathrm{pCT}}(\mathbf{x}) - \bar{D}_f(\mathbf{x}) \right)^2 }, \quad
\bar{D}_f(\mathbf{x}) = \frac{1}{K} \sum_{k=1}^{K} D_f^{(k) \rightarrow \mathrm{pCT}}(\mathbf{x}).
\end{equation}

Each fraction $f$ is then assigned a weight \( w_f \), reflecting the degree of anatomical variability or confidence in the session. 

\subsection*{Final Uncertainty Aggregation Across Models}
The final voxel-wise accumulated uncertainty up to fraction $F$ is computed as a weighted aggregation of the per-fraction uncertainties $\sigma_f(\mathbf{x})$, with $f \in \{1,\dots,F\}$ indexing each treatment session:
\begin{equation}
\sigma_{\text{acc}}^F(\mathbf{x})=
\frac{1}{\sum_{f=1}^{F} w_f}
\sqrt{\sum_{f=1}^{F}\left(w_f\,\sigma_f(\mathbf{x})\right)^2}.
\end{equation}

Here, $w_f$ represents the anatomical-variability–based weight assigned to fraction $f$. This formulation performs a weighted quadratic aggregation of model-driven registration uncertainty, where the uncertainty is first estimated independently for each fraction and then combined across all fractions up to the final session $F$. By aggregating uncertainty progressively at the fraction level, this approach avoids estimating uncertainty solely from the final accumulated dose, thereby reducing potential compounding effects.

To quantify the impact on dose–volume histograms (DVHs), confidence intervals are derived using $\pm 3\sigma_{\text{acc}}^F$. Under a Gaussian assumption, the $\pm 3\sigma$ range corresponds to an approximate 99.7\% coverage, which is a standard conservative choice in UQ.

\subsection{Evaluation Protocol}

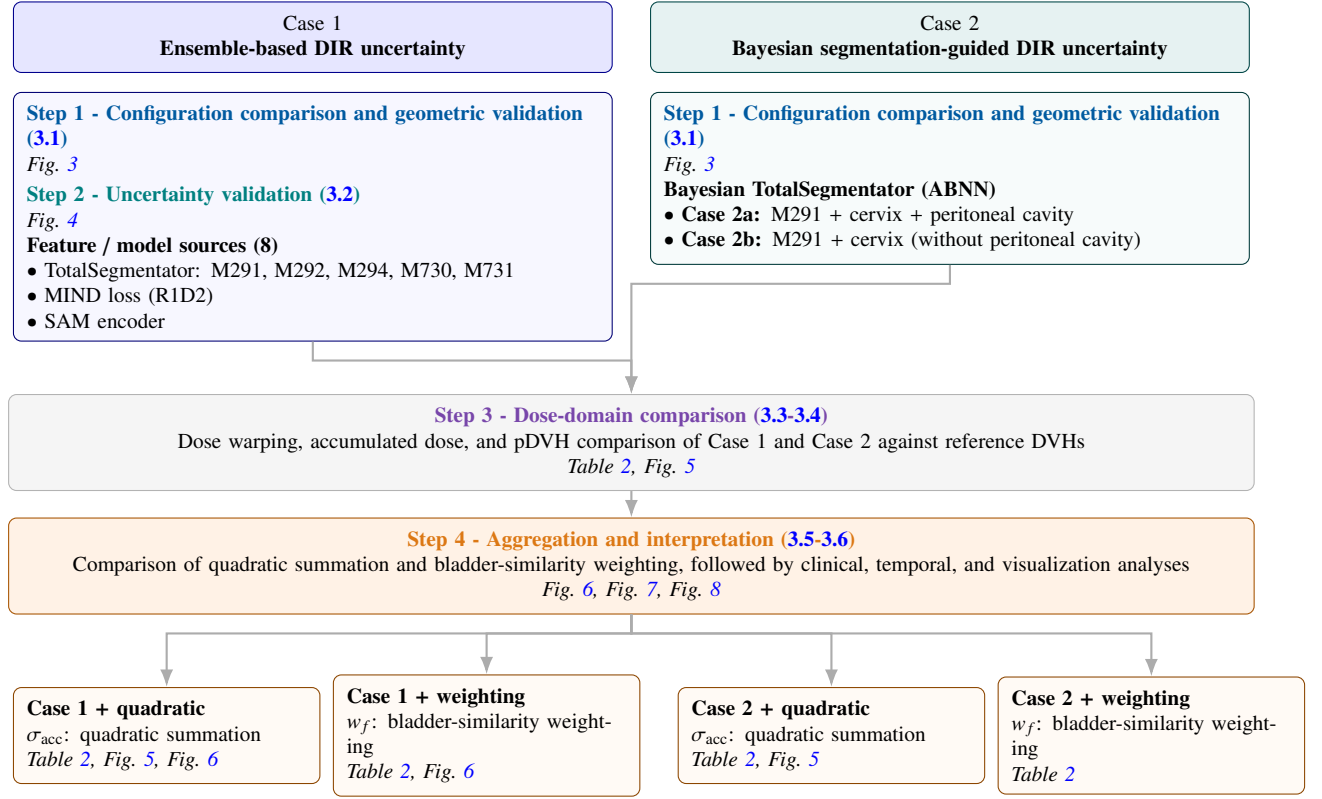
\begin{figure}[t]
\centering
\setlength{\tabcolsep}{0pt}
\renewcommand{\arraystretch}{1.05}

\begin{tikzpicture}[
  font=\footnotesize,
  roadmap/.style={draw=gray!60, rounded corners=3pt, fill=gray!12, align=center, inner sep=5pt, text width=#1},
  casehead1/.style={draw=blue!55!black, rounded corners=3pt, fill=blue!10, align=center, inner sep=5pt, text width=#1},
  casehead2/.style={draw=teal!55!black, rounded corners=3pt, fill=teal!10, align=center, inner sep=5pt, text width=#1},
  casebox1/.style={draw=blue!45!black, rounded corners=3pt, fill=blue!3, align=left, inner sep=5pt, text width=#1},
  casebox2/.style={draw=teal!45!black, rounded corners=3pt, fill=teal!3, align=left, inner sep=5pt, text width=#1},
  bridge/.style={draw=gray!60, rounded corners=3pt, fill=gray!8, align=center, inner sep=5pt, text width=#1},
  agghead/.style={draw=orange!65!black, rounded corners=3pt, fill=orange!10, align=center, inner sep=5pt, text width=#1},
  strategy/.style={draw=orange!55!black, rounded corners=3pt, fill=orange!4, align=left, inner sep=5pt, text width=#1},
  arrow/.style={-Latex, line width=0.8pt, draw=gray!65}
]

\coordinate (LEFT) at (0,0);

\def\W{0.46\linewidth}
\def\WA{0.225\linewidth}
\def\Gap{0.03\linewidth}

\node[casehead1=\W, anchor=north west] (c1h) 
{Case 1\\\textbf{Ensemble-based DIR uncertainty}};

\node[casehead2=\W, right=\Gap of c1h] (c2h)
{Case 2\\\textbf{Bayesian segmentation-guided DIR uncertainty}};

\node[casebox1=\W, below=2.5mm of c1h] (c1) {
\StepOne{\textbf{Step 1 - Configuration comparison and geometric validation (\ref{result_geometric})}}\\
\textit{Fig.~\ref{fig:dice_boxplot}}\\[2pt]
\StepTwo{\textbf{Step 2 - Uncertainty validation (\ref{result_correlation})}}\\
\textit{Fig.~\ref{fig:asd_scatterplot}}\\
\textbf{Feature / model sources (8)}\\
\(\bullet\) TotalSegmentator: M291, M292, M294, M730, M731\\
\(\bullet\) MIND loss (R1D2)\\
\(\bullet\) SAM encoder\\[2pt]
};

\node[casebox2=\W, below=2.5mm of c2h] (c2) {
\StepOne{\textbf{Step 1 - Configuration comparison and geometric validation (\ref{result_geometric})}}\\
\textit{Fig.~\ref{fig:dice_boxplot}}\\
\textbf{Bayesian TotalSegmentator (ABNN)}\\
\(\bullet\) \textbf{Case 2a:} M291 + cervix + peritoneal cavity\\
\(\bullet\) \textbf{Case 2b:} M291 + cervix (without peritoneal cavity)\\[2pt]
};

\node[bridge=0.98\linewidth, below=12mm of $(c1.south)!0.5!(c2.south)$] (dose)
{\StepThree{\textbf{Step 3 - Dose-domain comparison (\ref{result_coverage}-\ref{result_dvh})}}
\\
Dose warping, accumulated dose, and pDVH comparison of Case~1 and Case~2 against reference DVHs
\\
\textit{Table~\ref{tab:dvh_coverage}, Fig.~\ref{fig:dvh_1_2}}
};

\node[agghead=0.98\linewidth, below=3.5mm of dose] (agh)
{\StepFour{\textbf{Step 4 - Aggregation and interpretation (\ref{result_aggregation}-\ref{result_temporal})}}
\\
Comparison of quadratic summation and bladder-similarity weighting, followed by clinical, temporal, and visualization analyses
\\
\textit{Fig.~\ref{fig:dvh_case2_weighted}, Fig.~\ref{fig:dvh_clinical_goal}, Fig.~\ref{fig:temporal_uncertainty}}
};

\coordinate (a11pos) at ($(c1.west |- agh.south) + (0,-16mm)$);

\node[strategy=\WA, anchor=west] (a11) at (a11pos) {
\textbf{Case 1 + quadratic}\\
\(\sigma_{\text{acc}}\): quadratic summation\\
\textit{Table~\ref{tab:dvh_coverage}, Fig.~\ref{fig:dvh_1_2}, Fig.~\ref{fig:dvh_case2_weighted}}
};

\node[strategy=\WA, right=0.01\linewidth of a11] (a12) {
\textbf{Case 1 + weighting}\\
\(w_f\): bladder-similarity weighting\\
\textit{Table~\ref{tab:dvh_coverage}, Fig.~\ref{fig:dvh_case2_weighted}}
};

\node[strategy=\WA, right=0.03\linewidth of a12] (a21) {
\textbf{Case 2 + quadratic}\\
\(\sigma_{\text{acc}}\): quadratic summation\\
\textit{Table~\ref{tab:dvh_coverage}, Fig.~\ref{fig:dvh_1_2}}
};

\node[strategy=\WA, right=0.01\linewidth of a21] (a22) {
\textbf{Case 2 + weighting}\\
\(w_f\): bladder-similarity weighting\\
\textit{Table~\ref{tab:dvh_coverage}}
};

\draw[arrow] (c1.south) -- ++(0,-2.5mm) -| (dose.north);
\draw[arrow] (c2.south) -- ++(0,-2.5mm) -| (dose.north);
\draw[arrow] (dose.south) -- (agh.north);
\draw[arrow] (agh.south) -- ++(0,-2.5mm) -| (a11.north);
\draw[arrow] (agh.south) -- ++(0,-2.5mm) -| (a12.north);
\draw[arrow] (agh.south) -- ++(0,-2.5mm) -| (a21.north);
\draw[arrow] (agh.south) -- ++(0,-2.5mm) -| (a22.north);

\end{tikzpicture}
\caption{Roadmap of the investigation pipeline and evaluated uncertainty configurations. Case~1 estimates DIR uncertainty using an ensemble of heterogeneous feature sources, whereas Case~2 uses a Bayesian segmentation-guided strategy (variants 2a and 2b). \StepOne{Step~1} compares all candidate configurations geometrically (Subsection~\ref{result_geometric}). \StepTwo{Step~2} then evaluates whether the ensemble-based uncertainty estimate of Case~1 is spatially consistent with ASD-based geometric error (Subsection~\ref{result_correlation}). \StepThree{Step~3} compares Case~1 and Case~2 in the dose domain using accumulated dose and pDVHs against reference DVHs (Subsections~\ref{result_coverage} and \ref{result_dvh}). \StepFour{Step~4} compares quadratic and bladder-similarity weighted aggregation and uses the selected strategy for clinical, temporal/spatial, and visualization-oriented interpretation (Subsections~\ref{result_aggregation}--\ref{result_temporal}). 
\\ Abbreviations: DIR, deformable image registration; ASD, average surface distance; pDVH, probabilistic dose–volume histogram; ABNN, adaptable Bayesian neural network.}
\label{fig:cases_aggregation_overview}
\end{figure}

\hl{Figure~\ref{fig:cases_aggregation_overview} provides a roadmap of the investigation pipeline and links each step to the corresponding Results subsections, while the evaluation protocol follows this four-step structure from configuration comparison to clinically interpretable dose endpoints.} Dosimetric evaluation (Figure~\ref{fig:cases_aggregation_overview}) of the oART strategy was performed on the CTVt and on OARs: bladder and rectum.

\hl{Step 1 focused on configuration comparison and geometric validation. Geometric accuracy was assessed independently from dose accumulation to verify that the deformation fields were anatomically consistent.} Geometric accuracy of each registration method (Case~1, Case~2a, and Case~2b), together with the individual feature sources used in Case~1, was evaluated using contour overlap and surface-distance metrics.

\hl{Step 2 evaluated whether the ensemble-based uncertainty estimate could be interpreted as a spatial indicator of registration error.} To assess the reliability of the uncertainty estimated by Case~1, we performed a correlation analysis between voxel-wise uncertainty maps and the \hl{average surface distance (ASD)} computed between warped segmentations and ground truth contours. \hl{ASD was preferred here because it provides a continuous geometric error measure that is better suited for correlation analysis than DSC.} This evaluation determines whether regions of high estimated uncertainty systematically correspond to poor registration accuracy, thereby supporting the use of model disagreement as a surrogate for spatial registration confidence.

\hl{Step 3 compared Case~1 and Case~2 in the dose domain by propagating the estimated uncertainty to accumulated DVHs.} Accumulated DVHs were computed for Case~1 and Case~2, and compared with reference DVHs derived from summing the daily dose distributions recalculated on the anatomy of the day using ground truth contours. This comparison is necessary because no direct ground-truth accumulated dose is available in clinical practice. The daily-dose-based DVH therefore serves as the most accurate surrogate of the delivered dose, \hl{and the reported calibration should be interpreted relative to this surrogate rather than to an absolute ground-truth accumulated dose.}

The dose-domain comparison (Step~3 of Fig.~\ref{fig:cases_aggregation_overview}) includes:
\begin{itemize}
    \item calculation of the DVH coverage of each Case, defined as the percentage of the reference DVH falling within the predicted uncertainty interval;
    \item use of uncertainty amplitude across Cases to ensure a fair comparison of coverage performance;
    \item representative patient-wise pDVH comparisons between Cases.
\end{itemize}

\hl{Step 4 then focused on aggregation and interpretation. The aggregation analysis evaluated whether anatomical-variability weighting provided a more efficient uncertainty interval than quadratic summation, by comparing DVH coverage and interval amplitude across strategies. Overall, this step investigates how different uncertainty propagation strategies translate into clinically meaningful dosimetric differences.}

Based on these results, Case 1 (ensemble-based) yielding the highest coverage and best dosimetric agreement was selected for in-depth analysis.

\hl{The selected strategy was then further examined through clinical-goal evaluation, temporal slice visualization, and dedicated visualization tools in order to support spatial and clinical interpretation.}


\section{Results}

\subsection{Configuration comparison and geometric validation}
\label{result_geometric}

\begin{figure}[h!]
    \centering
    \includegraphics[width=0.9\textwidth]{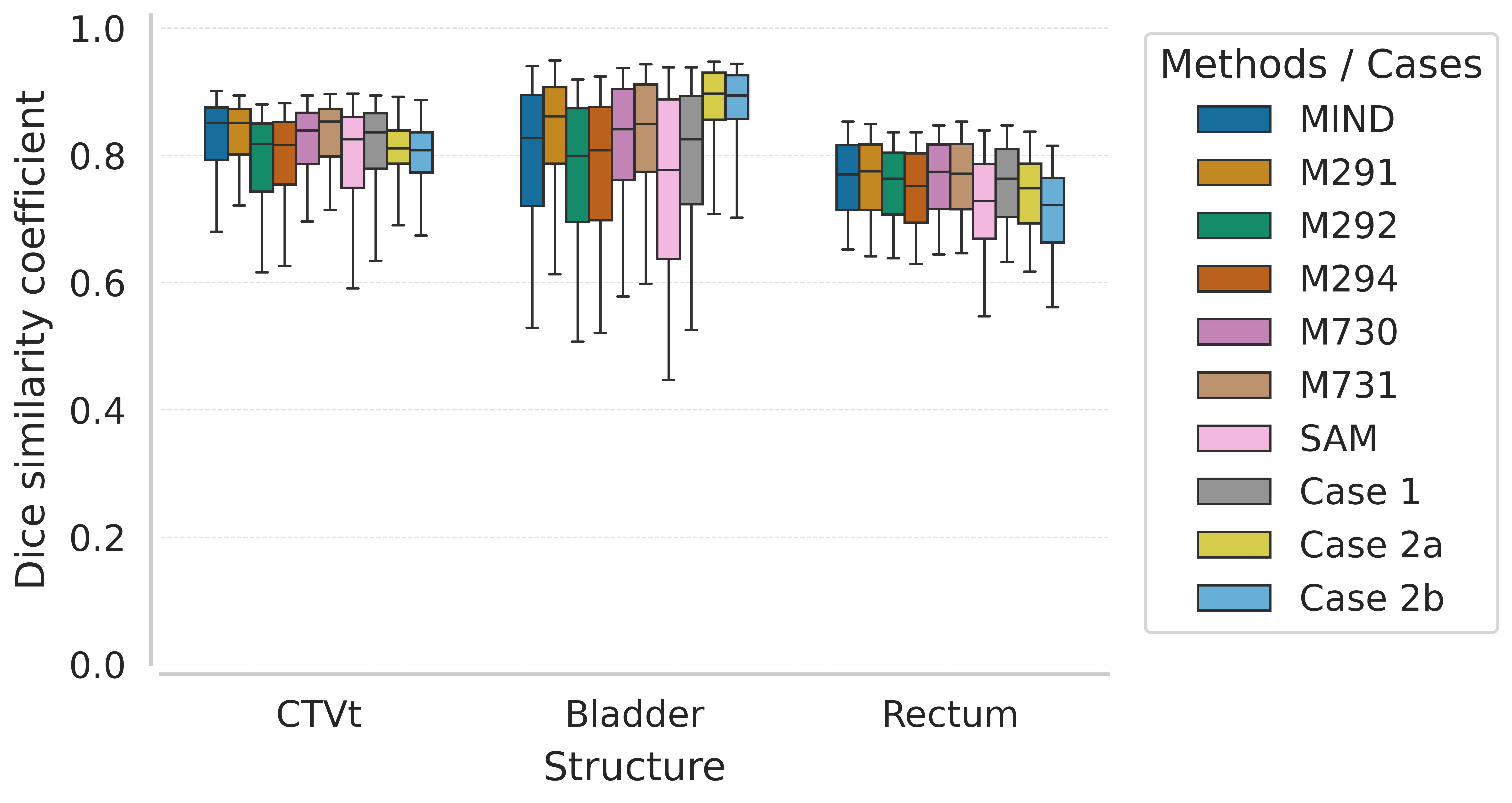}
    \caption{Boxplots of Dice similarity coefficients (DSC) between warped segmentations and clinician-validated ground truth contours for the CTVt, bladder, and rectum, across all patients. Results are shown for each individual model and for the three Cases: Case 1 (ensemble-based DIR), Case 2a (Bayesian segmentation–guided DIR with cervix and peritoneal cavity labels), and Case 2b (Bayesian DIR without peritoneal cavity label).}
    \label{fig:dice_boxplot}
\end{figure}

To evaluate geometric accuracy, DSC values were computed between warped segmentations (CBCTs) and ground truth contours (pCT) across all patients and structures. Figure~\ref{fig:dice_boxplot} presents boxplots of these results for each structure and across the different scenarios, including individual feature extraction models (e.g., MIND R1D2, SAM, TotalSegmentator variants) and the three predefined Cases (Case~1, Case~2a, and Case~2b). \hl{Additional exploratory analyses using ASD and HD95 showed similar overall trends across methods.}

For the CTVt, both the individual models and the ensemble-based strategy (Case 1) achieved the highest median DSC values, generally above 0.8, reflecting robust alignment for this relatively deformable target. In contrast, the Bayesian strategies (Case 2a and Case 2b) showed consistently lower performance for the CTVt.

For the bladder, DSCs varied more widely across methods, with noticeable variability for SAM and TotalSeg model M292/M294. For the rectum, all methods demonstrated lower DSCs, consistent with its deformable anatomy and less well-defined tissue boundaries. Overall, Case 1 provided more stable and consistent geometric accuracy across all three structures compared with the probabilistic segmentation-based registration approaches (Case 2a, 2b).

\subsection{Correlation between registration uncertainty and geometric error}
\label{result_correlation}

\begin{figure}[h!]
    \centering
    \includegraphics[width=1.025\textwidth]{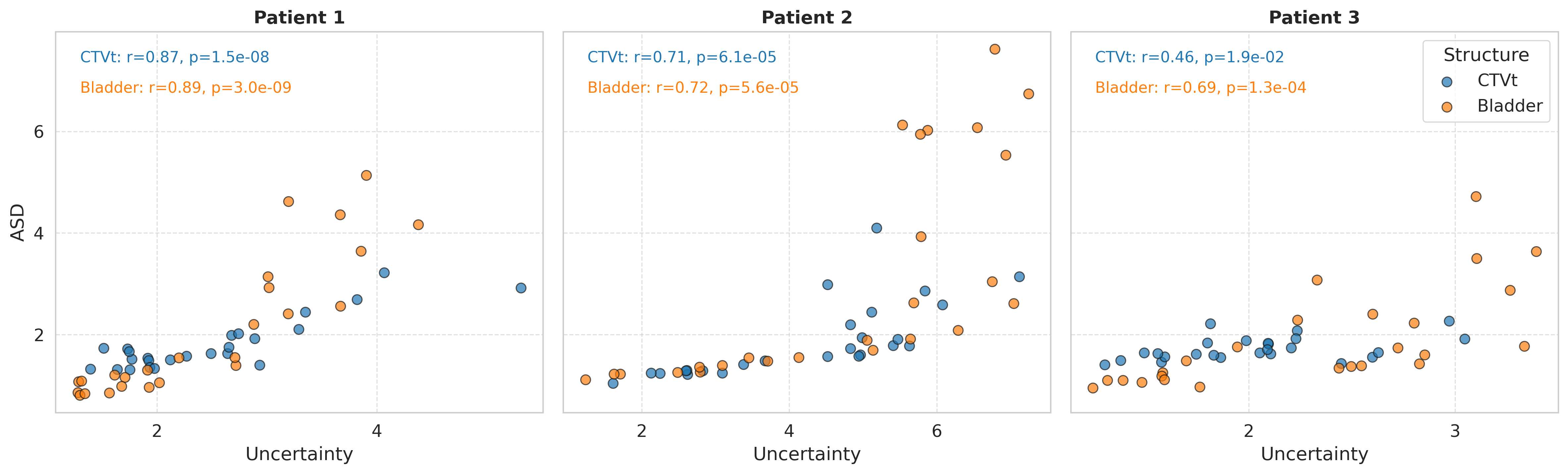}
    \caption{Scatterplots illustrate the relationship between contour registration uncertainty (x-axis) and average surface distance (ASD) to ground truth contours (y-axis) for the CTVt and bladder in three patients for Case~1. The rectum is not displayed because no consistent correlation pattern was observed. Registration uncertainty is defined as the 95th percentile of voxel-wise standard deviation across the ensemble-generated DVFs (Case 1), computed within a 5 mm dilation of each structure. Pearson correlation coefficients (r) between uncertainty and ASD are provided.}
    \label{fig:asd_scatterplot}
\end{figure}

To assess the relationship between registration uncertainty and spatial alignment error, we evaluated the correlation between voxel-wise uncertainty and the average surface distance computed between warped structures (obtained using the mean DVF) and ground truth contours. For each patient and each organ, the analysis was restricted to a 5 mm isotropic dilation of the structure to focus on regions most sensitive to registration errors. The uncertainty value for each structure was defined as the 95th percentile of the voxel-wise standard deviation across the deformation fields generated by the ensemble-based registration strategy (Case 1). This percentile-based approach highlights the most uncertain regions while reducing the influence of noise in low-gradient areas (regions of uniform intensity). Because the rectum exhibited no consistent correlation pattern and its inclusion hindered visual readability, only the CTVt and bladder are shown in Figure~\ref{fig:asd_scatterplot}.

As shown in Figure~\ref{fig:asd_scatterplot} for 3 patients representing the range of observed correlation strengths (low, median, and high), strong correlations between the uncertainty and the contour registration quality were observed for the CTVt and bladder, with mean Pearson correlation coefficients across the 9 patients of $0.628 \pm 0.192$ and $0.662 \pm  0.184$, respectively. In contrast, the rectum exhibited a weak correlation (mean coefficient $< 0.3$).

\subsection{Dose-domain comparison: dose–volume histograms coverage and calibration}
\label{result_coverage}

Table~\ref{tab:dvh_coverage} summarizes the percentage of the reference DVHs covered by the predicted uncertainty intervals (\(\pm3\sigma\)) for each aggregation strategy, together with the mean uncertainty amplitude. Case 1 with anatomical weighting achieved the highest CTVt coverage (96.3 $\pm 3.9\%$). For the bladder and rectum, coverage reached $84.8 \pm 20.8\%$ and $76.2 \pm 18.5\%$, respectively, performing better than the alternative strategies for similar uncertainty magnitudes.

Table~\ref{tab:dvh_coverage} summarizes the percentage of the reference DVHs covered by the predicted uncertainty intervals (\(\pm3\sigma\)) for each aggregation strategy, together with the mean uncertainty amplitude. Case 1 with anatomical weighting achieved the highest CTVt coverage (($96.3 \pm 3.9\%$)). For the bladder, the highest coverage was obtained with Case 2 weighting (($88.6 \pm 16.0\%$)). For the rectum, Case 1 with anatomical weighting provided the best coverage (($76.2 \pm 18.5\%$)), while maintaining comparable uncertainty magnitude.

The bold values in Table~\ref{tab:dvh_coverage} indicate the highest coverage achieved for each organ within comparable uncertainty amplitudes.

\begin{table}[ht]
\centering
\caption{Coverage of reference DVHs within the uncertainty interval (mean $\pm$ standard deviation) and associated uncertainty magnitude (Gy), for each structure and aggregation strategy. Bold values indicate the highest coverage achieved for a given structure.}
\label{tab:dvh_coverage}
\begin{tabular}{lcc cc cc}
\toprule
& \multicolumn{2}{c}{\textbf{CTVt}} 
& \multicolumn{2}{c}{\textbf{Bladder}} 
& \multicolumn{2}{c}{\textbf{Rectum}} \\
\cmidrule(r){2-3} \cmidrule(r){4-5} \cmidrule(r){6-7}
\textbf{Case} 
& Coverage (\%) & Unc. (Gy) 
& Coverage (\%) & Unc. (Gy) 
& Coverage (\%) & Unc. (Gy) \\
\midrule
Case 1 (quadratic)       
& 90.37 ± 8.27  & 0.27 
& 81.88 ± 20.69 & 1.09 
& 74.21 ± 20.76 & 0.76 \\

Case 1 (weighting)      
& \textbf{96.32 ± 3.85}  & 0.27 
& 84.81 ± 20.80 & 1.09 
& \textbf{76.15 ± 18.45} & 0.76 \\

Case 2 (quadratic)       
& 78.33 ± 12.24 & 0.25 
& 88.05 ± 17.25 & 1.25 
& 71.13 ± 21.39 & 0.93 \\

Case 2 (weighting)      
& 80.58 ± 11.35 & 0.25 
& \textbf{88.57 ± 15.96} & 1.25 
& 71.80 ± 22.63 & 0.93 \\
\bottomrule
\end{tabular}
\end{table}

\subsection{Dose-domain comparison: representative probabilistic DVHs}
\label{result_dvh}

\begin{figure}[ht]
   \centering
  \includegraphics[width=1\textwidth]{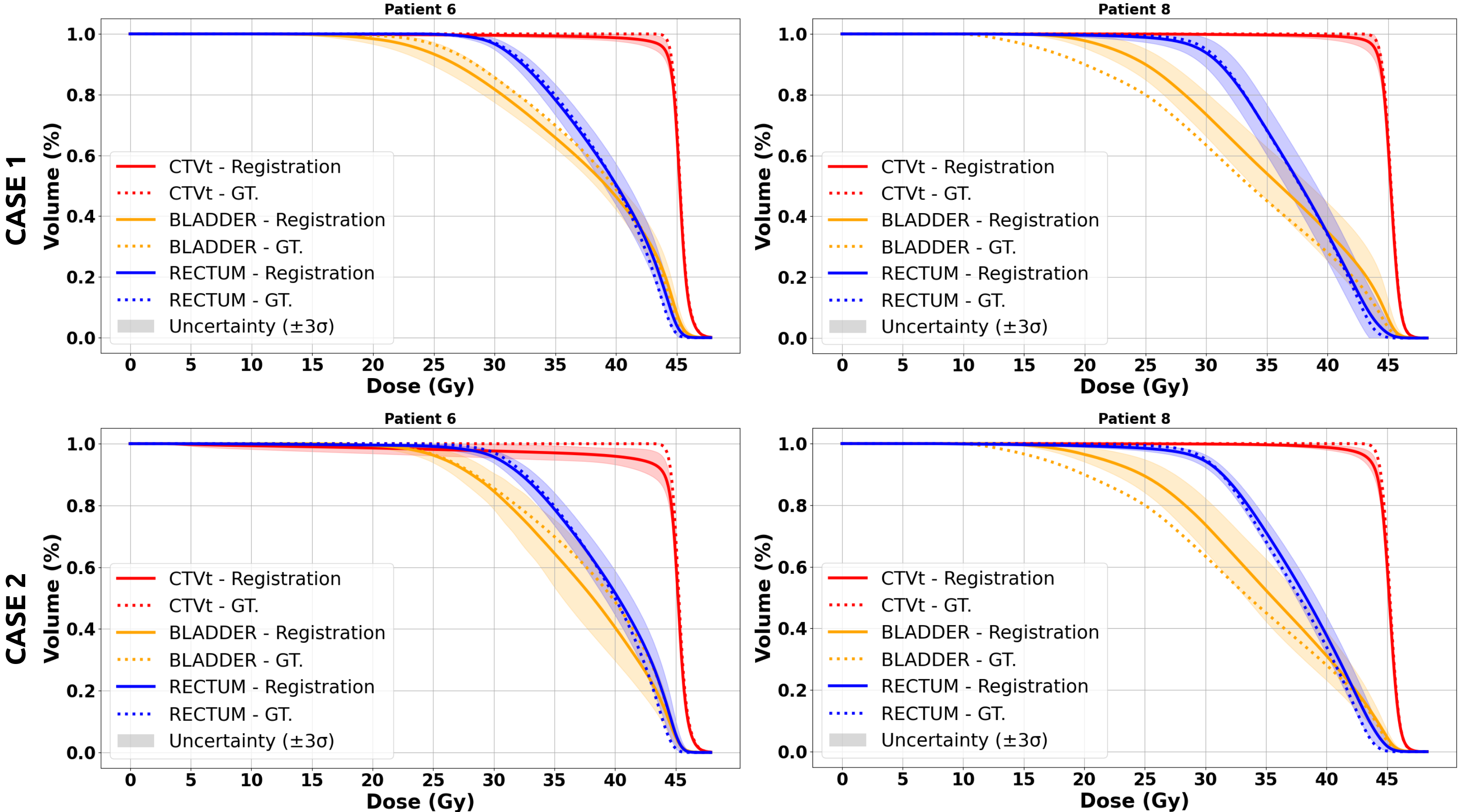}     
    \caption{Comparison of accumulated DVHs for Case 1 and Case 2 (quadratic) with respect to the ground truth (GT) for two representative patients. Patient 6 corresponds to the best-case scenario (highest DVH coverage), whereas Patient 8 represents the worst-case scenario (lowest DVH coverage). The left column displays Patient 6, showing the largest proportion of reference DVH points falling within the $\pm 3\sigma$ uncertainty band (transparent region), while the right column shows Patient 8, corresponding to the lowest coverage.}
  \label{fig:dvh_1_2}
\end{figure}

Figure~\ref{fig:dvh_1_2} illustrates the accumulated DVHs for two representative patients, comparing the performance of Case 1 and Case 2 against reference DVHs derived from daily doses using manual segmentations. \hl{Patient 6 represents the best-case scenario, defined as the highest coverage for the bladder and CTVt, whereas Patient 8 corresponds to the lowest coverage.} Case 1 consistently aligns more closely with the reference DVH, particularly in the high-dose region of the CTVt, highlighting the improved accuracy and robustness of the ensemble-based method. In contrast, Case 2 exhibits greater variability, including both under- and overestimation, especially for organs at risk.

\subsection{Aggregation strategy comparison and clinical interpretation}
\label{result_aggregation}

\begin{figure}[ht]
    \centering
    \includegraphics[width=1\textwidth]{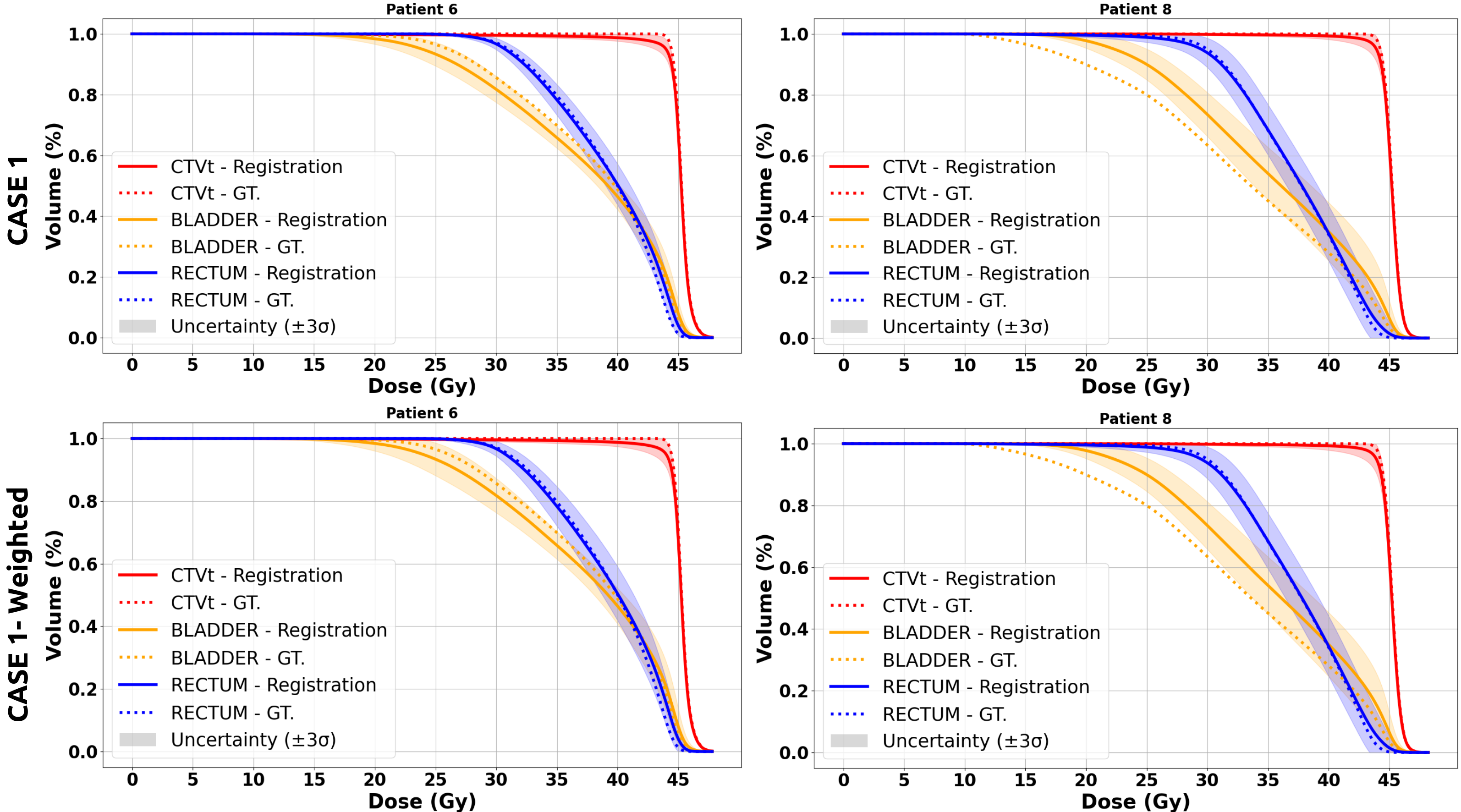}
   \caption{Comparison of two uncertainty aggregation strategies for Case 1 (ensemble-based): quadratic summation versus Case 1 weighted (anatomical-variability weighting). Both provide similar accuracy for the CTVt, but the weighted strategy produces noticeably narrower uncertainty intervals for the bladder and rectum, illustrating the benefit of incorporating daily anatomical variability.}
   \label{fig:dvh_case2_weighted}
\end{figure}

Figure~\ref{fig:dvh_case2_weighted} compares two uncertainty summation strategies applied to Case 1: quadratic summation versus anatomical variability–based weighting. While both achieve high accuracy for the CTVt, the weighted strategy results in narrower uncertainty intervals, particularly for the bladder and rectum. This demonstrates the benefit of accounting for anatomical changes in uncertainty modeling, leading to better-calibrated probabilistic DVHs.

\begin{figure}[ht]
    \centering
    \includegraphics[width=1\textwidth]{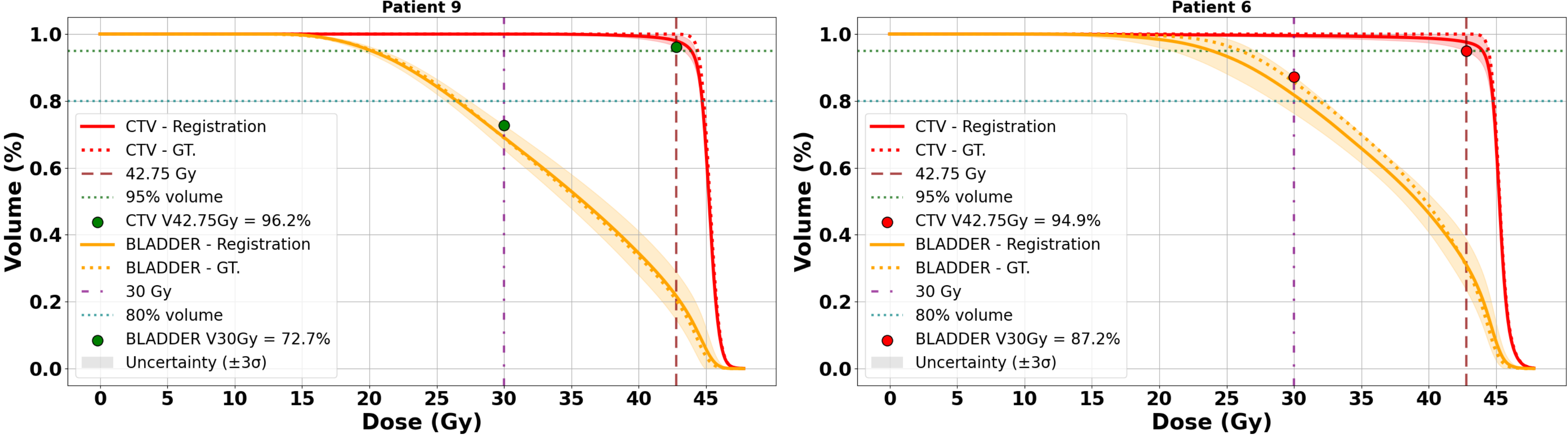}
    \caption{Illustration of how probabilistic DVHs (Case 1) support clinical decision-making. For two representative patients, accumulated DVHs are shown together with their $\pm3\sigma$ uncertainty intervals. The clinical constraints evaluated are bladder $V30Gy < 80\%$ and CTVt $V42.75Gy \geq 95\%$. Vertical and horizontal dashed lines indicate the corresponding dose and volume thresholds. The upper uncertainty bound allows identification of potential violation of organ-at-risk constraints (bladder), while the lower uncertainty bound may indicate insufficient target coverage (CTVt). This visualization provides a safety margin beyond mean DVH evaluation.}
    \label{fig:dvh_clinical_goal}
\end{figure}

The propagation of dose uncertainty can provide direct support for clinical decision-making. Beyond quantifying deviations from the planned dose, the probabilistic DVHs enable clinicians to evaluate whether predefined clinical goals are robustly satisfied under uncertainty. As illustrated in Figure~\ref{fig:dvh_clinical_goal}, the clinical constraints bladder $V_{30\mathrm{Gy}} < 80\%$ and CTVt $V_{42.75\mathrm{Gy}} \geq 95\%$ can be evaluated not only on the mean DVH but also by incorporating the propagated uncertainty intervals ($\pm3\sigma$). The dashed vertical and horizontal lines indicate the dose and volume thresholds for each structure, enabling direct visual assessment of constraint robustness.

For the bladder, exceeding the upper uncertainty bound beyond the 80\% volume threshold at 30 Gy indicates a potential violation under plausible registration scenarios. Conversely, for the CTVt, a decrease of the lower uncertainty bound below the 95\% volume threshold at 42.75 Gy signals a possible underdosage. In this framework, uncertainty acts as a safety margin, allowing clinicians to identify situations where mean-based evaluation may either underestimate toxicity risk or overestimate target coverage.

\subsection{Temporal and spatial interpretation of uncertainty}
\label{result_temporal}

\begin{figure}[ht]
    \centering
    \includegraphics[width=1\textwidth]{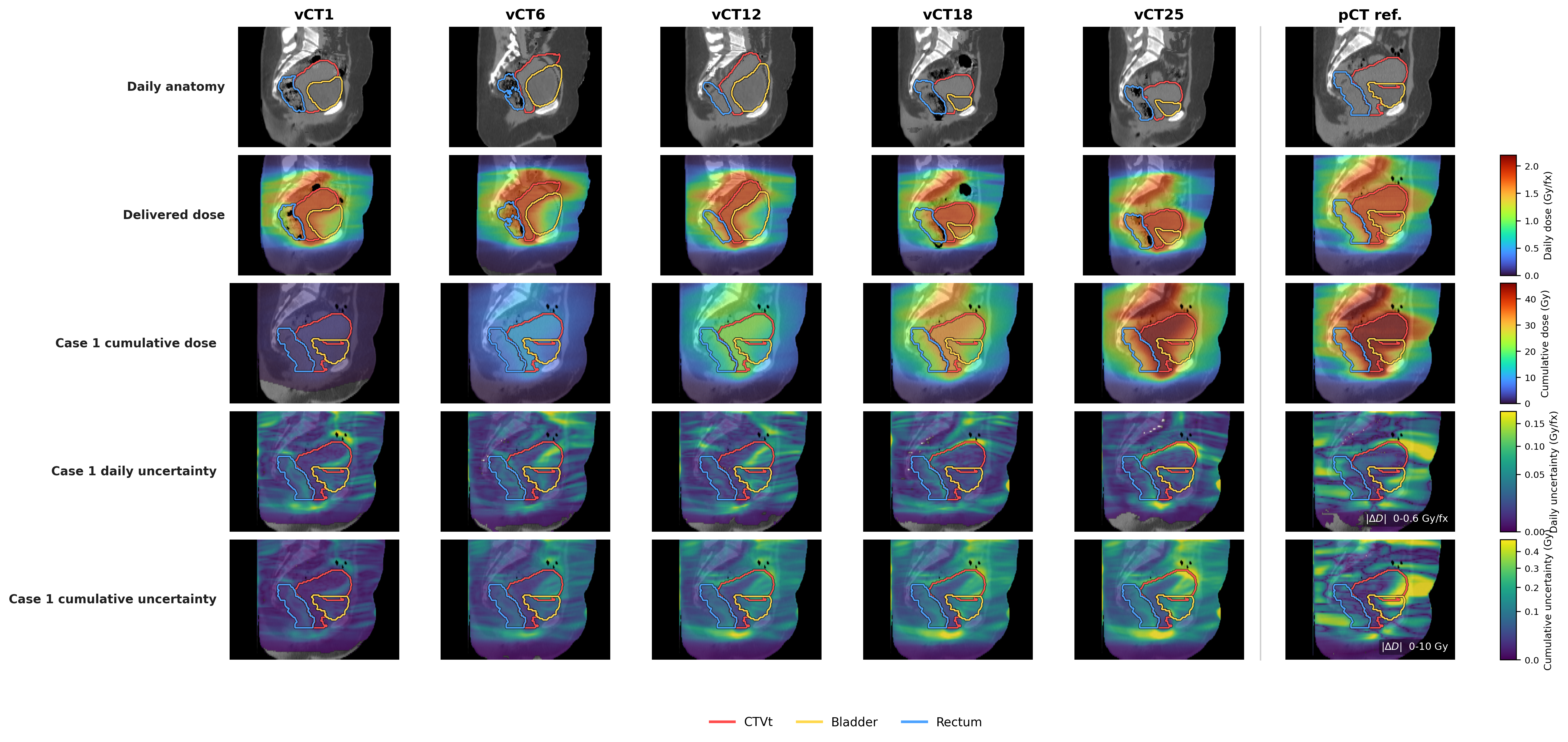}
    \caption{Temporal evolution of Case~1 dose propagation and uncertainty for Patient~4. Columns 1 to 5 show selected treatment fractions (vCT$1$, vCT$6$, vCT$12$, vCT$18$, and vCT$25$), and the last column shows the pCT reference. Row 1: native vCT anatomy with daily CTVt, bladder, and rectum contours. Row 2: delivered daily dose on the native vCT. Row 3: cumulative mean dose mapped to the planning CT reference frame. Row 4: daily dose uncertainty for the selected fraction, quantified as the voxel-wise standard deviation across the Case~1 ensemble after deformation to the pCT. Row 5: cumulative dose uncertainty on the pCT, obtained by quadratic summation of the fraction-wise standard deviation maps up to the displayed fraction. In the pCT reference column, rows 4 and 5 display the absolute dose difference \(|\Delta D|\) relative to the planning dose, allowing qualitative comparison between uncertainty localization and dose deviation.}
    \label{fig:temporal_uncertainty}
\end{figure}

Variability across fractions was investigated because cervical anatomy undergoes substantial deformation throughout treatment.
\hl{Figure~\ref{fig:temporal_uncertainty} summarizes the temporal evolution of anatomy, delivered dose, and propagated Case~1 quadratic uncertainty on sagittal slices for a representative patient at selected fractions (vCT$1$, vCT$6$, vCT$12$, vCT$18$, and vCT$25$). The first row shows the native anatomy of each selected fraction together with the daily CTVt, bladder, and rectum contours, while the second row displays the corresponding delivered daily dose on the native vCT. Rows 3 to 5 are all shown in the common pCT reference frame in order to enable direct spatial comparison across fractions: cumulative mean dose, daily Case~1 dose uncertainty, and cumulative Case~1 dose uncertainty, respectively.}

This slice-based representation improves interpretability by linking the uncertainty maps directly to anatomical landmarks and clinically relevant structures. Daily uncertainty remains spatially heterogeneous, with higher values predominantly observed near soft-tissue interfaces and in regions where the CTVt is adjacent to highly deformable organs such as the bladder and rectum. These areas are particularly sensitive to inter-fraction anatomical changes and may therefore affect registration stability. As treatment progresses, the cumulative dose pattern progressively departs from the planned distribution, while cumulative uncertainty increases in similar anatomical regions. The final pCT reference column facilitates this comparison by showing the planning anatomy and planning dose, together with the absolute dose difference ($|\Delta D|$) relative to the planning dose, which qualitatively highlights the spatial correspondence between high-uncertainty areas and larger dose deviations.

\subsection{Dose Accumulation Uncertainty Visual Tools and Integration into 3DSlicer}
All components of the IMPACT ecosystem are implemented as modular plugins within 3DSlicer \cite{pieper20043d}. This open-source integration ensures seamless interoperability between registration, segmentation, synthesis, and dose accumulation tasks, while maintaining full transparency and reproducibility of the workflow. 
The IMPACT-DoseAcc module was developed using Python interfaces of 3DSlicer. It provides a user-friendly interface for performing uncertainty-aware dose accumulation directly from daily CBCT or vCT datasets. Users can visualize voxel-wise registration uncertainty and export uncertainty-weighted dose maps for clinical review or research analysis.

To ensure compatibility with diverse clinical pipelines, IMPACT-DoseAcc supports DICOM-RT import/export and uses the MRML scene architecture of 3DSlicer for efficient data management. The full plugin code and documentation are available as open-source under the Apache 2.0 license \href{https://github.com/vboussot/SlicerImpactDoseAcc}{https://github.com/vboussot/SlicerImpactDoseAcc}.

This integration bridges the gap between advanced uncertainty modeling and clinical usability, making IMPACT-DoseAcc readily deployable for research validation and translational studies in ART.

We also introduce pDVHs as a tool to quantify and communicate the uncertainty in clinical endpoints. Instead of a single deterministic DVH curve, pDVHs are represented with confidence intervals (e.g., mean ± 3$\sigma$), derived from the voxel-wise propagated uncertainties. These plots allow clinicians to assess how uncertainty may affect standard dosimetric criteria (e.g., D$_{95\%}$, V$_{30\mathrm{Gy}}$), and support more robust, uncertainty-aware evaluations of treatment quality.

Together, these tools provide both anatomical and dosimetric perspectives on the variability of the accumulated dose, and serve as a bridge between quantitative uncertainty estimation and clinical decision-making.

\section{Discussion}

This study presents an end-to-end framework for uncertainty-aware dose accumulation in CBCT-guided oART for cervical cancer, implemented as part of the IMPACT-DoseAcc module within the broader IMPACT ecosystem.
By combining ensemble-based DIR and Bayesian segmentation uncertainty, we demonstrate that registration uncertainty can be quantified and propagated through the full dose accumulation workflow.

IMPACT-DoseAcc extends the unified architecture introduced by IMPACT-Reg and IMPACT-Synth, enabling fully integrated UQ from image registration to cumulative dose evaluation. Through its implementation as a 3DSlicer plugin, the framework bridges methodological innovation and clinical usability, providing a reproducible and open-source platform for ART research and validation.

\subsection{Impact of uncertainty estimation on dose accumulation}

Our results show that ensemble uncertainty (Case 1) in the deformation fields correlates strongly with geometric error for the CTVt and bladder, supporting model variability as a practical surrogate for epistemic uncertainty in multimodal DIR (mean $r=0.66$ across patients). In contrast, the rectum exhibited weak correlations (mean $r<0.3$), consistent with its higher deformability and lower contrast boundaries, which makes the registration more sensitive to local anatomy, bowel gas, and contour variability. These observations are consistent with prior reports that anatomical change is often the main factor of dosimetric variability compared with the choice of DIR algorithm itself, as shown by Nenoff et~al.~\cite{nenoff_deformable_2020} in the context of pulmonary proton therapy, and help explain structure-specific disparities in uncertainty-error coupling.

Beyond correlation, we evaluated whether propagating DIR uncertainty improves the dose domain. Using voxel-wise uncertainty propagation, the proposed ensemble strategy (Case~1) with anatomical-variability weighting achieved higher pDVH coverage for the CTVt (96.3\%~$\pm$~3.9) with a small uncertainty amplitude (0.27~Gy) than Case~2, and offered competitive coverage for bladder and rectum (Table~\ref{tab:dvh_coverage}). Notably, anatomical-variability weighting (via bladder similarity) provided better-calibrated intervals than quadratic aggregation at similar amplitudes (Figure~\ref{fig:dvh_case2_weighted}), suggesting that simple, physiologically motivated weights help avoid over-conservatism while preserving reliability.

Positioned against the literature, our findings complement several strands of UQ work. Lowther et~al.~\cite{lowther_quantifying_2020} quantified inverse consistency errors (ICE) and showed high voxel consistency overall but highlighted clinically relevant uncertainties in OARs. Kainz et~al.~\cite{kainz_use_2022} used DVH overlays to expose summation errors due to ICE. Our pDVH coverage/calibration analysis extends these ideas from geometric surrogates to probabilistic dose endpoints, using per-voxel propagated uncertainty rather than ICE alone. In proton settings, Amstutz et~al.~\cite{amstutz_approach_2021,amstutz_quantification_2024} modeled DVF uncertainty with dose-gradient awareness, attaining $\pm$5\% voxel accuracy for large portions of the volume and showing that cumulative evaluation mitigates fraction-level fluctuations. We observe a similar benefit: anatomical-variability–weighted accumulation stabilizes uncertainty intervals without inflating their amplitude, particularly for the bladder, as evidenced by the narrower confidence bands and maintained DVH coverage in Fig.\ref{fig:dvh_case2_weighted} and Table \ref{tab:dvh_coverage}.

Deep learning–based UQ methods (Smolders et~al.~\cite{smolders_deep_2023}) have reported algorithm-agnostic uncertainty prediction for lung and H\&N; our approach is synergistic but distinct in two ways: 
(\textit{i}) it is evaluated in a CBCT$\rightarrow$CT oART workflow where multimodal imaging amplifies uncertainty sources, and (\textit{ii}) it propagates voxel-wise uncertainty to accumulated dose and pDVHs, enabling confidence statements on clinical constraints. 
Together, these results suggest that uncertainty-aware accumulation can bridge geometric UQ metrics and clinically interpretable dose confidence.

\subsection{Clinical relevance and adaptive implications}

Clinically, uncertainty propagation changes how accumulated dose information is interpreted. Rather than treating a single deterministic DVH as a definitive outcome, clinicians can evaluate whether the probabilistic DVH satisfies key clinical constraints such as bladder V$_{30\mathrm{Gy}}<80\%$ or CTVt V$_{42.75\mathrm{Gy}}\geq95\%$ coverage (Fig.~\ref{fig:dvh_clinical_goal}). If the upper bound exceeds an OAR constraint, or if the lower bound falls below a target-coverage constraint, this indicates a potential violation under plausible registration scenarios and justifies a more conservative response, such as plan review, reoptimization, or closer patient monitoring. This approach builds on previous DVH-overlay QA studies~\cite{kainz_use_2022} and the Delta Index framework~\cite{van2024quantifying}, extending them by providing quantified confidence intervals directly within the DVH space.

Our findings provide practical guidance for adaptive clinical decision-making. First, the strong correlation between registration uncertainty and geometric error (Fig.\ref{fig:asd_scatterplot}) for the CTVt and bladder supports structure-specific adaptive schemes. For example, treatment re-optimization can be considered when the upper limit of the bladder DVH approaches clinical constraints, whereas sustained rectal uncertainty in the presence of stable pDVHs may prompt additional imaging or contour revision.

Second, because cumulative dose estimation tends to average out stochastic fluctuations between fractions (as also reported by Amstutz et~al.~\cite{amstutz_quantification_2024}), persistent upper-bound exceedances across several sessions may serve as a more reliable trigger for adaptation than single-fraction outliers.

Finally, uncertainty-aware evaluation criteria offer new possibilities for margin management and risk stratification. When upper bounds approach CTVt underdose thresholds, a transient margin expansion or plan library change may be justified; conversely, consistently narrow intervals may confirm the margin choice. Overall, integrating voxel-wise uncertainty into the accumulated dose transforms adaptive decision-making from a heuristic threshold to a confidence-weighted assessment.

\subsection{Methodological considerations and limitations}

Despite encouraging results, several limitations must be acknowledged. \hl{The relatively small cohort size (nine patients across four centers) restricts statistical power and generalizability. This study should therefore be interpreted as a proof-of-concept evaluation on longitudinal clinical data, and the reported correlations and pDVH calibration metrics may remain sensitive to outliers and patient-specific effects.} The calibration of the uncertainty intervals ($\pm3\sigma$) could not be empirically validated due to the limited number of ensemble models used in this proof-of-concept implementation. Increasing the ensemble size would likely improve statistical robustness, but at a higher computational cost.  

In addition, dose computations were performed on vCTs derived from pCT data, which ensured geometric consistency across fractions but did not fully represent the true daily anatomy. \hl{These vCTs provide a pragmatic surrogate for anatomy-of-the-day dose recalculation, but they may still contain geometric inaccuracies, are not independently validated against a ground-truth daily CT, and may be correlated with the DVFs used during dose accumulation. The reported calibration should therefore be interpreted primarily as relative consistency with respect to this surrogate rather than as absolute ground-truth dosimetric accuracy.} 

\hl{Finally, the multi-institutional nature of the dataset introduces additional heterogeneity in CT/CBCT systems, acquisition protocols, and clinical workflows. Although the use of vCTs partly harmonizes the intensity domain, no center-specific analysis was performed, and part of the observed performance may therefore still be influenced by institution-dependent factors.}

The quality of vCTs (strongly dependent on DIR performance and delineation accuracy) also remains a potential source of variability in dose recalculation. Finally, the use of a reduced 5~mm isotropic margin, though supported by prior studies~\cite{wang2024assessing,yen2022spare,kuipers2024margin}, may limit direct comparisons with non-adaptive strategies typically employing 10 to 25~mm margins \cite{Lim2011-marges}. Expanding the Bayesian segmentation framework to include additional pelvic structures could further refine anatomical priors and enhance the robustness of future uncertainty modeling.

\subsection{Perspectives and future work}

A comprehensive uncertainty-aware RT workflow should integrate all major uncertainty sources, from image acquisition to dose accumulation. The prospective combination of CBCT-based dose computation, whether derived from CBCT, MRI, or other imaging modalities, together with Bayesian segmentation and ensemble-based DIR, will enable a fully probabilistic characterization of treatment delivery. This framework is also compatible with sCTs generated using state-of-the-art deep learning approaches~\cite{huijben2024generating}, which may further enhance anatomical fidelity and enable comprehensive uncertainty quantification across both image synthesis and dose computation processes. 

\hl{Beyond methodological developments, future efforts should focus on the integration of such uncertainty-aware tools into clinical workflows. In particular, embedding these approaches within commercial treatment planning systems would facilitate their adoption by clinicians and support routine use in adaptive radiotherapy. Finally, although this study focuses on cervical cancer, the proposed framework is designed to be generic and could be extended to other anatomical sites with different motion patterns and clinical constraints.}

Ultimately, this uncertainty propagation framework contributes to the broader effort of establishing trust in AI-driven ART \cite{lambert2024trustworthy,huet2024quantifying}. Quantifying and visualizing uncertainty remain essential steps toward bridging the gap between algorithmic performance and clinical safety.

\section{Compliance with Ethical Standards}
\label{sec:ethics}
This research was conducted in accordance with statutory requirements. It has been approved by the regional research ethics committee (Comité de Protection des Personnes, Ouest V, Rennes, ARCOL 2015-02-45-01). All participants gave written informed consent to participate in the study.

\section{Acknowledgments}
\label{sec:acknowledgments}
The present work was funded by a PhD scholarship Grant from Elekta AB and by the French National Research Agency as part of the DIMADOSE project (C.Hémon), by the Ligue Contre le Cancer and the French National Cancer Institute (INCa) as part of a national clinical research grant (PHRC ARCOL). While preparing this work, the authors used ChatGPT to enhance the writing structure and refine grammar. After using these tools, the authors reviewed and edited the content as needed and took full responsibility for the publication’s content.

\bibliography{biblio.bib}
\bibliographystyle{ieeetr} 

\appendix
\section{Patient-wise comparison of weighted pDVHs}

\hl{Figure~\ref{fig:appendix_dvh_case1_case2_weighted} in the Appendix provides a patient-wise comparison of weighted pDVHs obtained with Case~1 and Case~2. Each row corresponds to one patient, with Case~1 weighted shown on the left and Case~2 weighted on the right. The CTVt, bladder, and rectum curves are displayed together with the corresponding reference DVHs, and the colored shaded regions indicate the associated $\pm 3\sigma$ intervals.}

\begin{figure}[p]
    \centering
    \includegraphics[width=1\textwidth,height=0.93\textheight,keepaspectratio]{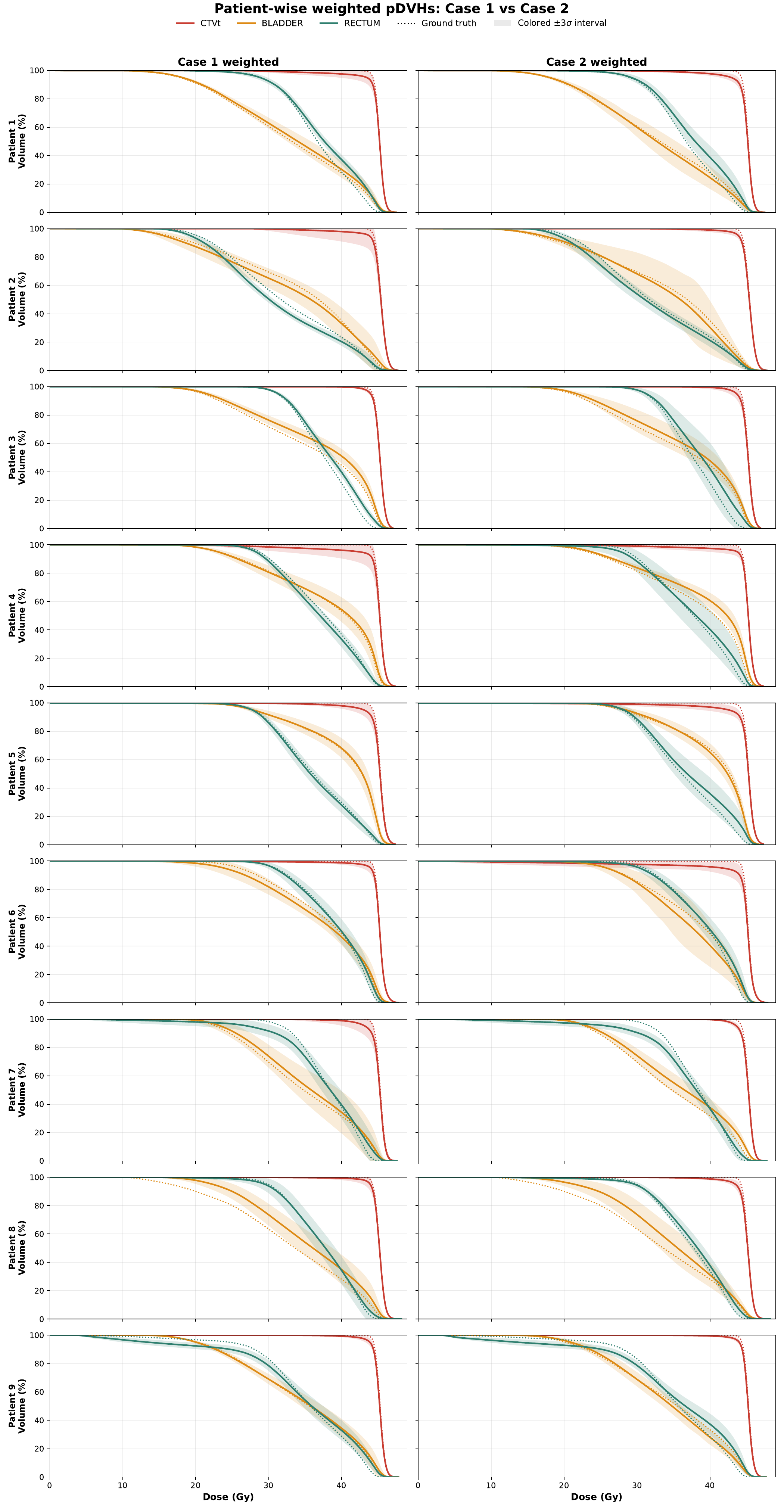}
    \caption{Patient-wise comparison of weighted pDVHs for Case~1 and Case~2. Each row corresponds to one patient, with Case~1 weighted shown on the left and Case~2 weighted on the right. Red curves denote the CTVt, orange curves denote the bladder, and green curves denote the rectum. Dotted lines correspond to the ground truth, and the colored shaded regions indicate the associated $\pm 3\sigma$ intervals.}
    \label{fig:appendix_dvh_case1_case2_weighted}
\end{figure}

\end{document}